\def\year{2021}\relax
\documentclass[letterpaper]{article} 
\usepackage{aaai21}  
\usepackage{times}  
\usepackage{helvet} 
\usepackage{courier}  
\usepackage[hyphens]{url}  
\usepackage{graphicx} 
\usepackage{algorithm}
\usepackage{algorithmic}

\usepackage{natbib}  
\usepackage{caption} 
\usepackage{float}
\usepackage[switch]{lineno}

\usepackage{multirow}
\usepackage[dvipsnames]{xcolor}
\usepackage{amsmath} 
\usepackage[utf8]{inputenc}
\usepackage{amssymb}
\usepackage{hyperref}
\usepackage{xcolor}

\providecommand{\keywords}[1]{\textbf{\textit{Keywords ---}} #1}

 \nocopyright 
\makeatletter
\newcommand{\settitle}{\@maketitle}
\makeatother

\title{Semi-Supervised Learning for Multi-Task Scene Understanding by \\ Neural Graph Consensus}
\author {
Marius Leordeanu\textsuperscript{1,2}\quad
Mihai Pirvu\textsuperscript{1}\quad
Dragos Costea\textsuperscript{1}\\
Alina Marcu\textsuperscript{1,2}\quad
Emil Slusanschi\textsuperscript{1}\quad
Rahul Sukthankar\textsuperscript{3}\\
}
\affiliations {
    \textsuperscript{\rm 1} University Politehnica of Bucharest \\
    \textsuperscript{\rm 2} Institute of Mathematics of the Romanian Academy \\
    \textsuperscript{\rm 3} Google Research\\
    \email{\{marius.leordeanu,mihai\_cristian.pirvu,dragos.costea,alina.marcu,emil.slusanschi\}@upb.ro}
    \email{sukthankar@google.com}
}

\begin{document}
\maketitle


\begin{abstract}
We address the challenging problem of semi-supervised learning in the context of multiple visual interpretations of the world by finding consensus in a graph of neural networks. Each graph node is a scene interpretation layer, while each edge is a deep net that transforms one layer at one node into another from a different node. During the supervised phase edge networks are trained independently. During the next unsupervised stage edge nets are trained on the pseudo-ground truth provided by consensus among multiple paths that reach the nets' start and end nodes. These paths act as ensemble teachers for any given edge and strong consensus is used for high-confidence supervisory signal. The unsupervised learning process is repeated over several generations, in which each edge becomes a "student" and also part of different ensemble "teachers" for training other students. By optimizing such consensus between different paths, the graph reaches consistency and robustness over multiple interpretations and generations, in the face of unknown labels. We give theoretical justifications of the proposed idea and validate it on a large dataset. We show how prediction of different representations such as depth, semantic segmentation, surface normals and pose from RGB input could be effectively learned through self-supervised consensus in our graph. We also compare to state-of-the-art methods for multi-task and semi-supervised learning and show superior performance.
\end{abstract}


\section{Introduction}

We propose the \textbf{Neural Graph Consensus (NGC)} model, a multi-class graph of deep neural networks, which approaches one of the most difficult problems in AI, that of unsupervised learning of multiple scene interpretations. Space-time data is so inexpensive to record, yet so expensive to annotate. While classic deep learning is powerful, it is almost hopeless when no ground truth is available. A popular alternative is reinforcement learning \cite{sutton1998introduction, silver2016mastering}, which is computationally very demanding when applied to real-world tasks. We show that NGC, based on a set of sound principles, overcomes many of the current limitations in unsupervised learning, by bringing together graphs and neural nets, within a single multi-task structure. We have ample evidence for the power of deep nets~\cite{ciresan2011flexible, krizhevsky2012imagenet, lecun2015deep}; given enough data and sufficient supervision they can master almost any task.
On the other hand, graphs are able to transcend to global solutions: starting from local knowledge they reach the global level through iterative message passing~\cite{pearl2014probabilistic,yedidia2005constructing,boykov2001fast,besag1986statistical}. Thus, iterative graph methods tend to converge to stable solutions, where consensus between the local and the global is reached and a higher ``truth'' emerges gradually. 
We take advantage of this property of classical graphical models and boost it by replacing the simple functions on nodes and edges, with powerful deep nets that will transform an entire scene representation into another. The idea also differs from current work in graph neural nets, which is mainly supervised and still used relatively simple functions on edges (e.g., MLPs) and local receptive fields~\cite{nicolicioiu2019recurrent, battaglia2016interaction, xu2018powerful}.

\begin{figure*} [!t]
\centering
\includegraphics[scale=0.435,keepaspectratio]{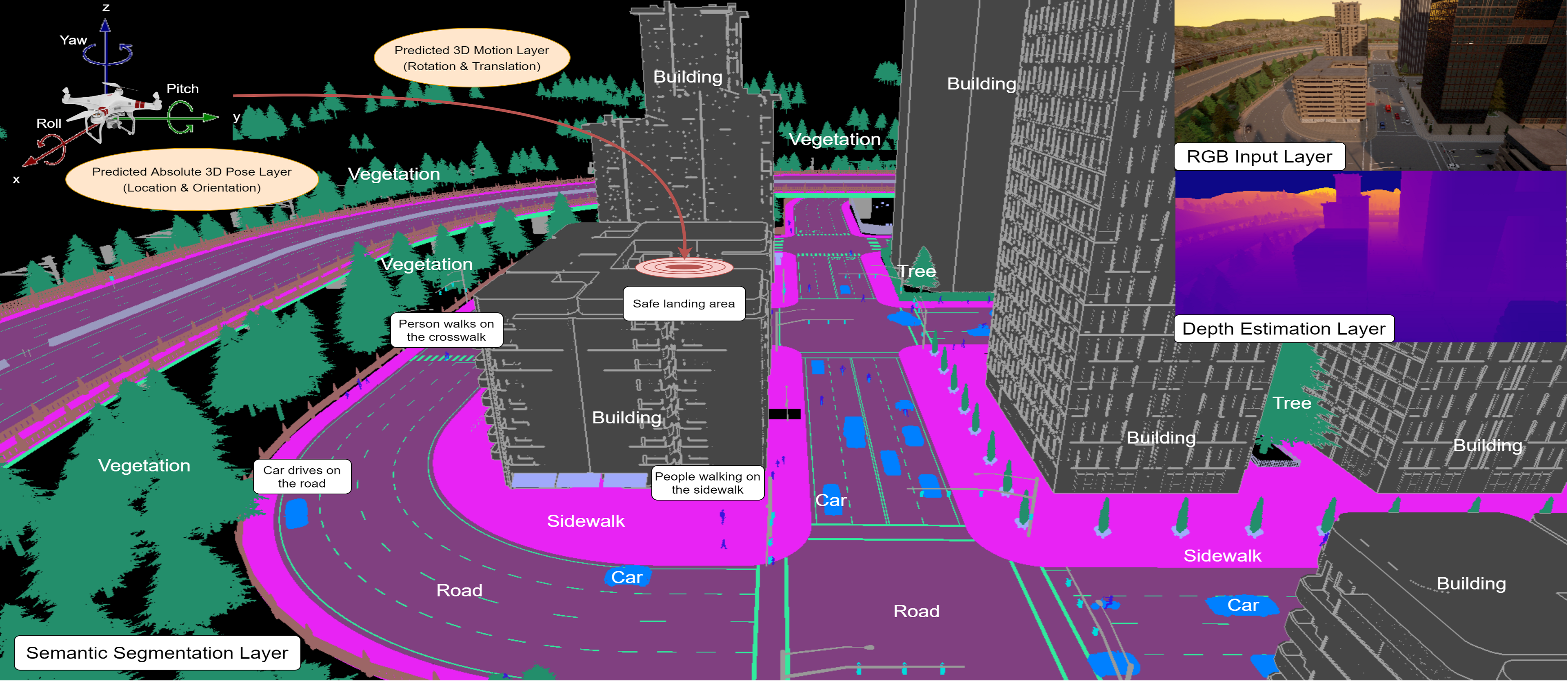}
\caption{\label{fig:GNC_Main_Concept} \textbf{NGC} can put together different interpretations of the dynamic scene, such as 3D structure, pose, motion, semantic segmentation of objects and activities in different regions of space and time, into a unified neural graph, in which multiple paths reaching a given node become teacher through consensual agreements to any single edge net reaching the same node. Trained in this self-supervised manner, NGC can reach robust unsupervised learning in the face of unlabeled data. The scene in the figure is taken from our virtual dataset, on which we run our experiments. While in this paper we do not take advantage of the temporal component, we show that NGC can effectively learn semi-supervised, to predict the drone 6D pose, scene depth, 3D structure, and semantic segmentation, from a single RGB image.}
\end{figure*} 

We create a graph (or more generally, a hypergraph) of deep networks, such that each node in the graph represents a different interpretation layer of the world (e.g. semantic segmentation layer, depth layer, motion layer). The edges (or hyperedges) have dedicated deep networks, which predict the layer at one node from the layer, or layers, at one or several other nodes. The deep nets are trained in turn, using as supervision the consensual output of all paths reaching the same output node (when such consensus takes place). Since we do not have strict, well defined ``worlds'' as in reinforcement learning (RL), we let the natural structure in the data emerge, as in graph clustering. Similar to RL, in which agents in heavily constrained environments with well defined goals evolve by playing against each other, we put pressure on the deep nets by training them one versus the others, in complementarity. Thus, each net brings something new, while also learning to agree with the other paths that reach the same output. During such consensus training, the NGC graph will approach a state of equilibrium. This idea is related to early work on grouping from the Gestalt school of psychology \cite{koffka2013principles} and Grossberg’s classical Adaptive Resonance Theory on how the brain learns and reaches conscious states 
\cite{grossberg1976adaptive, grossberg2000complementary, grossberg2015brain}. Interestingly, our idea also relates to the general concept of homeostasis~\cite{betts2016anatomy}, which is the innate tendency of all living organisms towards a relatively stable equilibrium between their many interdependent elements, in order to maintain their life.

In Fig.~\ref{fig:GNC_Main_Concept} we present the general NGC concept in the context of a drone that learns to understand its environment by finding consensus among the many different interpretations of the space-time world. Within a large but coherent NGC we see how different tasks, such as predicting semantic classes, 3D structure, pose, motion and detection of activities in different space-time regions, are inter-connected. 

\subsection{Scientific context}
There is current work that tackles unsupervised learning by constraining together multiple tasks, such as depth and relative pose~\cite{chen2019self, godard2019digging, zhou2017unsupervised, ranjan2019competitive, bian2019unsupervised, gordon2019depth, yang2018unsupervised}. Other works include semantic segmentation into the equation~\cite{tosi2020distilled, guizilini2020semantically, chen2019towards, stekovic2020casting}. There is also work that goes beyond vision, to consider input from other senses, for unsupervised learning by cross-modal prediction~\cite{Hu_2019_CVPR, Li_2019_CVPR, zhang2017split, pan2004automatic, he2017unsupervised, zhao2018sound}. These works are related to our NGC concept. By constraining different tasks, they are actually learning to satisfy consensus among multiple representation paths, even though they are limited to a few specific tasks with specific domain constraints.

There is also work that shows that clustering together several tasks to learn, through self-supervised agreement more compact features and descriptors could significantly help in learning, but the evaluation is still done in supervised settings. That is mainly because the unsupervised problem is kept at the level of unlabeled, anonymous clusters as it is usually done in the classic unsupervised clustering literature. Such papers include methods based on contrastive learning~\cite{chen2020simple,henaff2019data,caron2020unsupervised}. For example, Contrastive Multiview Coding~\cite{tian2019contrastive} proposes a compact descriptor from multiple representations. They show that a larger number of representations results in a better descriptor being learned for each scene. Another, related work, aims to maximize information from multiple views of a shared context~\cite{bachman2019learning}. A more general contrastive loss framefowrk is presented in \cite{patrick2020multi}. Learning from agreements among collaborative experts also proved useful in video retreival~\cite{liu2019use}.

An important unsupervised source of information is the temporal dimension which provides spatiotemporal consistency. This cue is exploited by an increasing number of recent papers for semantic segmentation propagation or pose propagation~\cite{marcu2020semantics, wang2019learning, jabri2020space} or interpretable keypoints~\cite{jakab2020self}. Our NGC model is general and can easily integrate information from different moments in time and take advantage of temporal consistency.

Multi-task learning research also starts taking advantage of consistency and agreements between multiple tasks: Taskology \cite{lu2020taskology} aims to do so by training pairs of tasks, using an unlabeled dataset to add a consistency loss term to the supervised one. Then, an idea that is even more related to ours, Cross-Task Consistency (CTC) \cite{zamir2020robust} aims to learn robust representations by using at least two intermediary representation conversions. It can be shown that their work could be seen as a simplified instance of our more general NGC model, but limited to fewer representations and not yet taking advantage of learning unsupervised over several iterations. Our NGC model is unique in the way it puts many well-defined tasks together, in a single hyper-graph structure, such that the output of one task can be the input of another. Then, NGC uses the consensual output among multiple pathways that transform one task into another and reach the same node (with a specific interpretation of the scene) 
as unsupervised, pseud-ground truth. After each iteration the models on the single task edges improve, overall consensus in the NGC hypergraph also improves, new data is added and the process continues.   

Therefore, instead of letting the final problem be supervised and confining unsupervised learning at the level of anonymous features (as most prior work does), our approach is in fact the opposite: put the many different final tasks inside the greater unsupervised learning graph system, then train them (in a classical way, using well-known costs published in the literature) on the unsupervised consensual output produced by the many pathways in the graph, which reach the input and output nodes corresponding to each task.

\begin{figure*}[!t]
\centering
\includegraphics[scale=0.57,keepaspectratio]{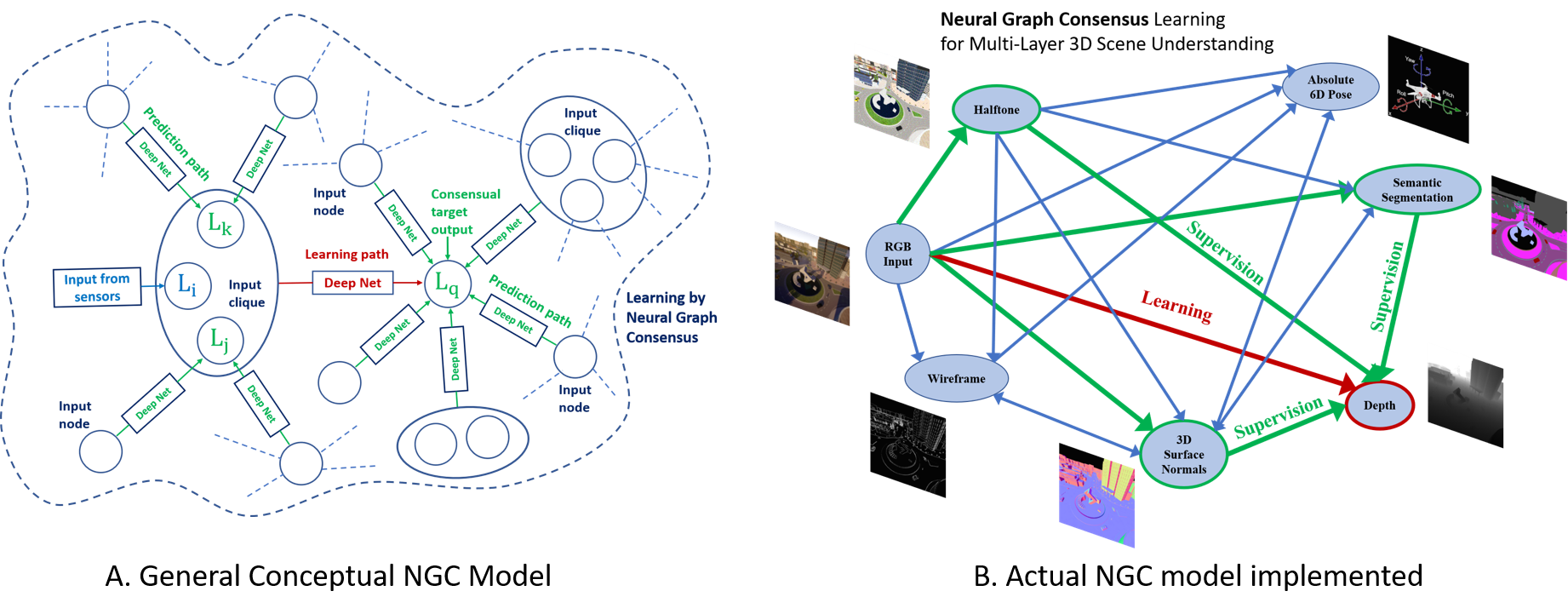}
\caption{\label{fig:NGC_Actual_Model} \textbf{A:} – formal representation of the NGC model as graph (or hypergraph) of deep nets. Each node has an interpretation layer of the “world”. Each edge learns to predict the layer at one node from one or (several) input layers from others. Each layer has an associated space-time region. NGC can operate in both space and time, with access to past input layers. The deep net shown in red, with input from a clique of layers learns to predict layer $L_q$, using as pseudo-ground truth, the state of $L_q$ reached by consensus by the green nets reaching node $q$. Layers at all nodes in NGC are either input from sensors or decided through consensus. The deep nets on edges take turns in being “trained students” or in “collaborating through consensus” as part of teacher ensembles, to train other nets. \textbf{B:} The actual NGC structure and representations used in the paper: learn semi-supervised to predict depth, semantic segmentation, absolute 6D pose (position and orientation) and 3D world structure, in the case of a drone flying in a virtual environment.}
\end{figure*} 

Our work is also related to classic ensemble approaches, which are very effective but limited to a single task. There is also work that shows how to use ensembles of nets for unsupervised learning \cite{croitoru2019unsupervised}. However, no other work shows how a single structure can put together a combinatorial number of ensembles within itself to learn simultaneously and semi-supervised many different tasks. Our results suggest that by putting all these ensembles and tasks together within a graph, we do not make the tasks harder, but easier, as the graph structure offers them access to a large number of teacher ensembles.

\subsection{Main contributions} 
We propose Neural Graph Consensus (NGC), a novel model for semi-supervised learning of multiple scene interpretations, which connects many deep nets in a single graph structure, with nets on edges and interpretation layers on nodes. Each net transforms one representation from one node into another, from a different node. We show how different tasks can teach themselves through self-supervised consensus. We offer theoretical justification of our approach and also prove its effectiveness in experiments and comparisons to top methods in the field.

\section{Neural Graph Consensus Model}

\begin{figure*} [!t]
\centering
\includegraphics[scale=0.555,keepaspectratio]{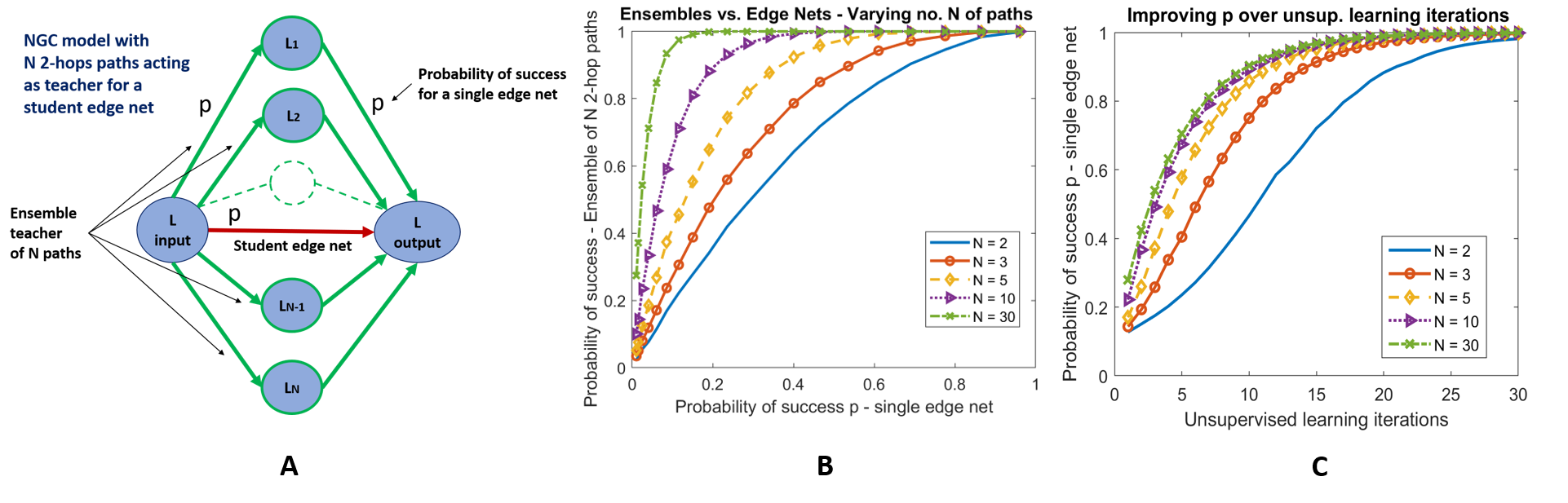}
\caption{\label{fig:simluation_plots} A. The simple NGC model with 2-hop pathways used in simulations. 
B. Performance of teacher ensemble vs.\ p, for different $N$ of ensemble paths and 100 classes per node. Plots are averages over 10k tests. C. Improving single edge nets over iterations, for different number of paths, when the next iteration edge nets recover $20\%$ of the performance gap to their teacher.}
\end{figure*} 

Each node $i$ in the graph (or hypergraph) has an associated layer $L_i$, which encodes a specific view or interpretation of the space-time world (e.g. depth map, semantic segmentation, regions in space-time for various activities). The layer at one node can be predicted from one or more other layers at other nodes by deep nets. These nets that transform one or multiple layers into another, form edges (or hyperedges) in the graph (hypergraph). In Fig.~\ref{fig:NGC_Actual_Model}A, we show the general concept: $L_i$, $L_j$ and $L_k$ form an input clique of nodes and $L_q$ is the layer at the output node. NGC can be arbitrarily connected: many paths starting from sensor inputs can reach the a given internal node.

\textbf{Unsupervised learning with NGC:} in general we expect to have access to limited labeled data, to pretrain separately a good part of our nets. Once we connect them into a NGC graph, we can start the unsupervised learning phase. Let us consider unsupervised training of the net associated with the hyperedge having input clique $(L_i, L_j, L_k)$ and output node $L_q$ (in red, Fig.~\ref{fig:simluation_plots}A). There could be many prediction paths (in green) going from other layers and sensors reaching $L_q$. The idea is that we let prediction information flow through the green paths and use their consensual outputs as supervisory signal for the red network. In our experiments we consider the median output as consensus (in the case of regression) and majority voting (in the case of classification). Note that all layers involved that are not coming from sensors (e.g. $L_i$, $L_j$, $L_k$,) are established by the same consensus strategy as $L_q$. Unsupervised learning will take turns: each net becomes student of the NGC (hyper)graph and is trained by the mutual consensus coming from the contextual pathways that reach the same output node. NGC becomes a democratic self-supervised system for which agreement becomes the ultimate teacher in the face of unlabeled data.

\subsection{Theoretical Analysis}
\label{sec:GNC_model_theory}

\textbf{Unsupervised learning in the case of regression:}
Let $L_q$ be a scalar value $x_q$. The case immediately extends to the
multivariate case. Let $x_q^{(c)}$  be the value at node $q$ predicted by the net coming from input clique $c$ to node $q$. Let $t_q$ be the ideal true value for node $q$. For ease of presentation, we consider L2 loss for training. In the supervised case, for the net going from input clique $c$ to output node $q$, the objective loss over all training cases is:	
$J_\mathrm{sup}\sum(x_q^c - t_q)^2$. In the unsupervised case, with many different paths reaching node $q$, we assume that the errors made by the predictive nets from all cliques $a$ to node $q$ are independent and unbiased. Then, we have $E_a(X_q^{(a)})=t_q$.
Also, based on the law of large numbers we expect that the ground truth can be approximated by 
$t_q = E_a(X_q^{(a)}) \approx \frac{1}{N_q}\sum_{q=1}^{N_q}x_q^{(a)}$,
in the small variance sense, as the number of paths going into node $q$ is large. Thus, in the unsupervised case we could use instead of $t_q$ its empirical approximation. Note that in practice we could do even better and use as pseudo-ground-truth, the median or a smart voting scheme. If we use the above approximation of $t_q$ in the L2 loss, the unsupervised loss $J_\mathrm{unsup}$ becomes:

\begin{equation}
\label{eq:unsup_loss}
    J_\mathrm{unsup} = \sum(x_q^c - \frac{1}{N_q}\sum_{q=1}^{N_q}x_q^{(a)})^2
\end{equation}

The unsupervised loss above is also an approximation of variance. Thus, by minimizing it we expect to minimize variance in outputs along different paths that reach the same node. This leads to the following conclusion:

\textbf{Proposition 1:} In a densely connected NGC graph, we expect the variance over the outputs reaching a given node to decrease during unsupervised learning. 

\textbf{Observation:} In our experiments over seven different tasks, in a single iteration of unsupervised learning the standard deviation over ensemble outputs reaching a specific node reduced nearly half ($56\% \pm 13\%$). This is empirical evidence that our proposed approach (Algorithm~\ref{alg:NGC_learning}) increases not just performance but also the level of overall agreement in the NGC graph.

 \begin{algorithm}[t!]
  \caption{Learning with Neural Graph Consensus}
 \label{alg:NGC_learning}
 \begin{algorithmic}
      \STATE \textbf{Step 1:} Pretrain a set of nets that transform different input to output representations, using the labeled data available.
      \STATE \textbf{Step 2:} Form the NGC graph by linking the nets such that the output of one (or several) becomes input to another. 
      \STATE \textbf{Step 3:} On a completely new unlabeled set, re-train the nets using as pseudo-ground truth for a specific node (representation) the consensual output of all paths that reach that node. Repeat Step 3, by choosing a new unlabeled set and newly trained nets, until convergence.
  \end{algorithmic}
  \end{algorithm}

\begin{figure*}[!t]
\centering
\includegraphics[scale=0.108,keepaspectratio]{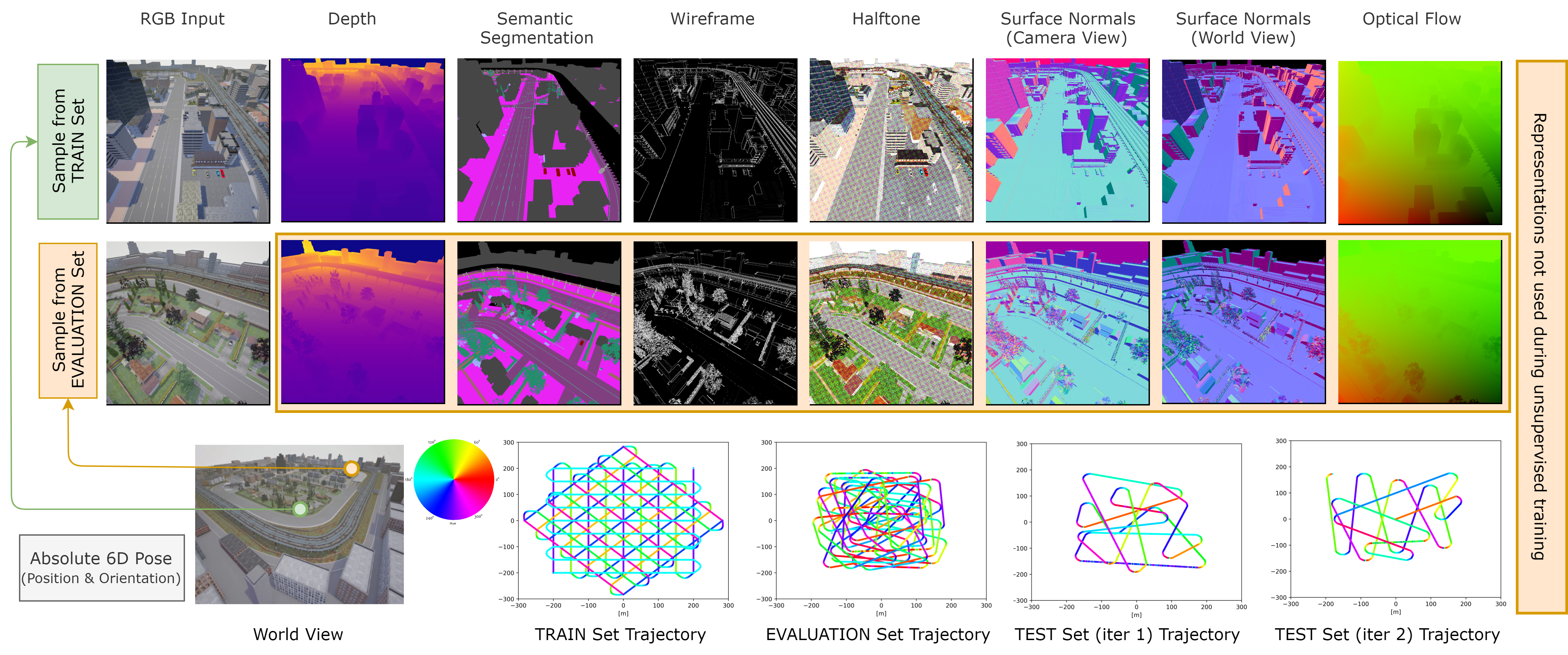}
\caption{\label{fig:dataset_main_figure} Samples from the synthetic dataset (train, first row, evaluation, second row). The hue for the trajectories (third row) encodes the yaw angle (see the color circle above). The paths simulate a drone trajectory, with small random variations in all angles. While the training path is a grid-based, traditional surveying flight, the test path aims to capture as much viewpoints as possible, for a robust evaluation. The distance between two adjacent training trajectories is 50m, chosen such that two parallel images have a small overlap.}
\end{figure*} 

\textbf{Unsupervised learning in the case of classification:}
Classification offers a more complex case: the agreements among paths are found by voting,
harder to analyze then simple averages. We consider the case with 2-hop paths that start from input and reach the same output node, after passing through some intermediate layer, to become teacher ensemble for the direct input-output link (Fig. \ref{fig:NGC_Actual_Model}A). For clarity of presentation we consider a single class per node, without loss of generality. At each node, a class is chosen by the incoming edge net. We assume that all edge nets function independently and the probability of success (correct class chosen given a specific input) is $p$. Then, what is the probability of success of the 2-hop teacher ensemble using majority voting? How does the number of classes $C$, the number $N$ of 2-hop paths and probability $p$ influence the performance of the ensemble vs. the single direct edge. This is important since the teacher provides the pseudo-ground truth for the next generation single edge nets (Alg. \ref{alg:NGC_learning}).

We make the following assumptions: for a given edge net, if the input class is correct than the probability of success is $p$, for a total number of $C$ classes. For a correct input, the chance of a wrong class at output is uniformly distributed among the remaining classes. For the first, input node, coming from sensors, we always consider its input as being correct. According to this model, we can show that the probability of success along any 2-hop pathway is $p_e^{(+)} = p^2 + (1-p)\frac{1}{C}$ and the probability of a wrong output over a 2-hop path is $p_e^{(-)}(1-p)(p+\frac{C-1}/C)$. We can show that $p_e^{(+)}$ for a single 2-hop path is still better than random chance $p_e > \frac{1}{c}$, if we assume $p > \frac{1}{c}$. 

Several key observations are relatively easy to prove: the percentage of correct votes $v_c$ over $N$ different 2-hop paths in an ensemble is expected to be $p_e^{(+)}$, while the expected percentage of wrong votes for any given wrong class $v_w$ is $p_e^{(-)}/(C-1)$. While the expected values do not depend on $N$, their variance decreases towards zero as $N \to \infty$, since: $\mathit{Var}(v_c) = \frac{p_e^{(+)}(1 - p_e^{(+)})}{N}$ and $\mathit{Var}(v_w)  \frac{p_e^{(-)}(1 - p_e^{(-)})}{N}$. Then we reach the following result: 

\textbf{Proposition 2:} If the success probability $p$ for an edge net is better than random $p > \frac{1}{C}$, then the probability of success $p_{eN}$ of the teacher ensemble (2-hop paths) using majority voting goes towards 1 as the number of paths $N \to \infty$. 

\noindent \textbf{Proof:}
Let
$\mu = p_e^{(+)} - p_e^{(-)}$
and
$\sigma = \sqrt{\mathit{Var}(v_c) + \mathit{Var}(v_w)}$
Then, the probability of error of an ensemble of $N$ 2-hop paths is (by Chebyshev's inequality): $p_{eN}^{(-)}=\mathit{Pr}(v_c - v_w < 0)\leq \mathit{Pr}(\|v_c - v_w - \mu\| < \mu) \leq \frac{\sigma^2}{\mu^2}$. If we plug in the formulas for $\sigma$, $\mu$, $\mathit{Var}(v_c)$ and $\mathit{Var}(v_w)$, we obtain:

\begin{equation}
   p_{eN}^{(-)} \leq \frac{1}{N} \frac{p_e^{(+)}(1 - p_e^{(+)}) + p_e^{(-)}(1 - p_e^{(-)})}{(p_e^{(+)} + p_e^{(-)})^2}  
\end{equation}

\begin{table*} [t!]
\caption{\label{table:table_representations_results} Results for our proposed ensemble NGC and distilled EdgeNets on 6 representations, over 2 iterations of unsupervised learning. We show best results over ensembles in NGC (\textcolor{red}{\textbf{bolded red}}) and single EdgeNets (\textcolor{blue}{\textbf{bolded blue}}. Note the consistent improvements from one iteration to the next for both ensembles teachers as well as single student nets.}
\begin{center}
\begin{tabular}{|l|l|c|c|c|c|c|}
\hline
& &  Iteration 0 & \multicolumn{2}{|c|}{Iteration 1} & \multicolumn{2}{|c|}{Iteration 2} \\
\cline{3-7}
Representation & Evaluation Metric & EdgeNet &	NGC & Distil. EdgeNet & NGC & Distil. EdgeNet\\	
\hline
  & L1 (meters) & 4.9844 & 3.4867 & 4.2802 & \textcolor{red}{\textbf{3.2994}} & \textcolor{blue}{\textbf{3.9508}}\\
\cline{2-7}
Depth & Pixels $\uparrow$ (\%) & - & 79.30 & 60.66 & \textcolor{red}{\textbf{79.69}} & \textcolor{blue}{\textbf{61.90}} \\
\cline{2-7}
\hline
Surface  & L1 (degrees) & 8.4862 & 7.7914 & 8.2891 & \textcolor{red}{\textbf{7.4503}} & \textcolor{blue}{\textbf{7.6773}} \\
\cline{2-7}
Normals (C) & Pixels $\uparrow$ (\%) & - & 74.18 & 53.59 & \textcolor{red}{\textbf{74.61}} & \textcolor{blue}{\textbf{53.94}} \\
\cline{2-7}
\hline
Surface & L1 (degrees) &  11.8859 & 8.8248 & 10.7500 & \textcolor{red}{\textbf{8.5282}} & \textcolor{blue}{\textbf{8.6714}}\\
\cline{2-7}
Normals (W) & Pixels $\uparrow$ (\%) & - & 79.95 & 57.88 & \textcolor{red}{\textbf{81.12}} & \textcolor{blue}{\textbf{61.14}}\\
\cline{2-7}
\hline
 & Accuracy & 0.9001 & 0.9181 & 0.9019 & \textcolor{red}{\textbf{0.9245}} & \textcolor{blue}{\textbf{0.9283}}\\
\cline{2-7}
Semantic Segmentation  & mIOU & 0.4840 & 0.4978 & 0.4980 & \textcolor{red}{\textbf{0.5258}} & \textcolor{blue}{\textbf{0.5159}}\\
\cline{2-7}
 & Pixels $\uparrow$ (\%) & - & 79.46 & 69.62 & \textcolor{red}{\textbf{81.49}} & \textcolor{blue}{\textbf{71.95}} \\
\cline{2-7}
\hline
 Wireframe & Accuracy & 0.9617 & 0.9655 & 0.9654 & \textcolor{red}{\textbf{0.9661}} & \textcolor{blue}{\textbf{0.9655}}\\
 \cline{2-7}
\cline{2-7}
 & Pixels $\uparrow$ (\%) & - & 77.71 & 72.57 & \textcolor{red}{\textbf{78.02}} & \textcolor{blue}{\textbf{73.46}}\\
\cline{2-7}
\hline
Position  & L2 (meters) & 25.7597 & 15.5383 & 20.0204 & \textcolor{red}{\textbf{12.0764}} & \textcolor{blue}{\textbf{15.5599}}\\
\hline
Orientation  & L1 (degrees) & 3.8439 & 2.5001 & 3.3961 & \textcolor{red}{\textbf{2.2088}} & \textcolor{blue}{\textbf{3.0005}}\\
\hline
\end{tabular}
\end{center}
\end{table*}

The result leads to the conclusion that the accuracy of the ensemble improves towards 1, as $N \to \infty$. We performed simulation experiments to visualize the accuracy of the teacher ensemble of 2-hop paths vs. the single edge nets, based on the model presented here (Fig. \ref{fig:simluation_plots}A and \ref{fig:simluation_plots}B). The improvement of the ensemble over the single net depends mostly on the number of paths. As predicted by the model, our simulations also verified that the number of classes $C$ does not influence significantly these curves (see appendix). We also simulated the effect of learning over multiple iterations. We consider the case when the single edge net of next iteration recovers only $20\%$ from the gap to the teacher, after being trained on the teacher output at the previous iteration (Fig. \ref{fig:simluation_plots}D). While our simulations are based on simplified mathematical models, they suggest an interesting intuitive observation: \textbf{if} 1) paths are many and independent (by having a large diverse pool of representations at nodes in NGC), and 2) single nets are able to keep up with their ensemble pathways over iterations, and 3) we keep bringing new unlabeled data from one iteration to the next (to maintain independence over time),
then we could expect continuous improvement over many iterations.

\section{Experimental analysis}
\label{sec:experiments}

\begin{figure*} [!t]
\centering
\includegraphics[scale=0.227,keepaspectratio]{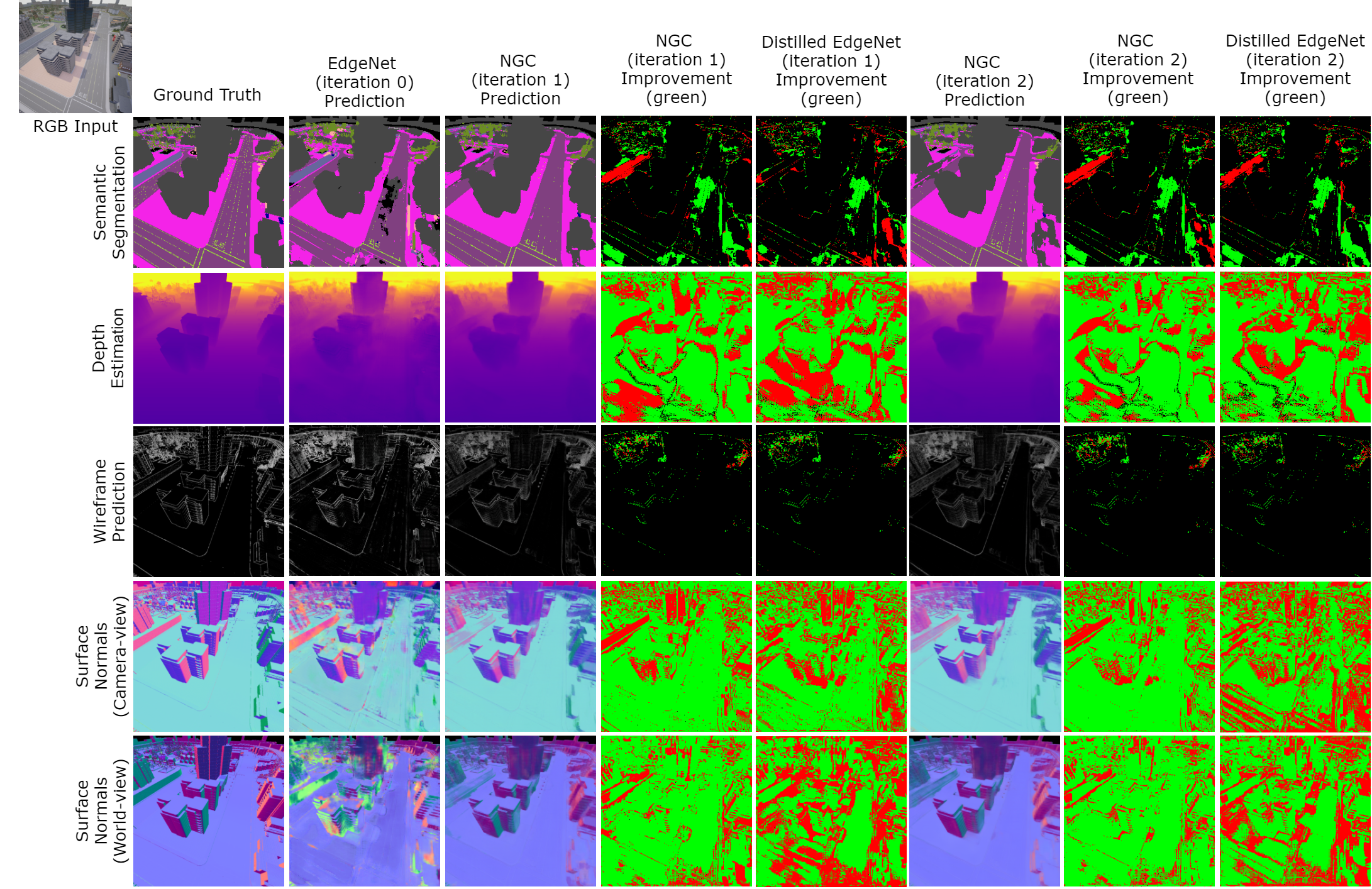}
\caption{\label{fig:improved_pixels_qual_results} Qualitative results for NGC. The plots show performance improvement (green) vs performance degradation (red) on five tasks. We compare the original predictions with two graph iterations and their distilled EdgeNets. }
\end{figure*} 

\textbf{Dataset details and description:}
To test the NGC approach in the case of many scene representations we capture a large dataset in a virtual environment (that we will publicly release), 
in which a drone flies above a city and learns to predict from a single image the scene
depth, the 3D surface normals (both from the world and camera system of reference), its absolute location and orientation (6D pose), the scene wireframe (object boundaries) as well as the semantic segmentation of the scene in 12 classes: building, fence, pedestrian, pole, road line, road, sidewalk, vegetation, vehicle, wall, traffic sign, other (Fig. \ref{fig:dataset_main_figure}). The dataset is divided in four: supervised training set (subdivided in 8k images for training and 2k for validation), 2 test sets (10 k images each, for unsupervised learning iterations 1 and 2) and a separate evaluation set (10 k images, never seen during learning). \\

\begin{table*}
\caption{\label{table:table_multitask_experiments} Multi-task learning results. All methods were trained on the same supervised data (our train set) and tested on the evaluation set. NGC uses the ensemble outputs of EdgeNet (iter 0). Single task networks (marked NDDR* below) were trained with and without pretraining (on other datasets than ours).}
\begin{center}
\begin{tabular}{|l|l|c|c|c|c|c|c|}
\hline
Task & Metric & EdgeNet(iter 0) & NGC & NDDR*(no pretrain) & NDDR*(pretrain) & NDDR & MTL-NAS\\
\hline
 Semantic & mIOU & 0.484 & \textbf{0.498} & 0.141 & 0.343 & 0.315 & 0.368\\
 \cline{2-8}
 Segm. & Acc. & 90.017 & \textbf{91.816} & 48.7 & 84.2 & 86.9 & 87.8 \\
\hline
Normals (C) & Err (deg.) & 8.4862 & 7.7914 & 9.820 & 7.727 & 6.801 & \textbf{6.533} \\
\hline
\end{tabular}
\end{center}
\end{table*}

\begin{table*}
\caption{\label{table:table_semi_experiments} Semantic segmentation comparisons on our evaluation set with the semi-supervised CCT\cite{ouali2020semi}. We outperform CCT both in absolute measures as well as relative improvement (semi-sup vs. supervised)}
\begin{center}
\begin{tabular}{|l|c|c|c|c|c|}
\hline
Metric & EdgeNet (iter 0) & NGC (iter 2) & EdgeNet (iter 2)  & CCT (supervised) & CCT (semi-sup) \\
\hline
 mIOU & 0.484 & \textbf{0.526} & 0.516 & 0.353 & 0.353 \\
 \cline{1-6}
  Accuracy  & 0.9001 & 0.9245 & \textbf{0.9283} & 0.8463 & \textbf{0.8503}\\
\hline
\end{tabular}
\end{center}
\end{table*}

\noindent \textbf{Implementation details:} We developed a general NGC framework on top of the existing deep learning framework PyTorch~\cite{NEURIPS2019_9015}, which can model arbitrary complex graphs and which we will make publicly available (Detailed in the appendix). For EdgeNets we used two types of networks: Map2Map and Map2Vector. The Map2Map architecture, used for most edge nets (since most output a image-size map) is based on a small UNet-style architecture (see appendix for details). The Map2Vector has the same encoder as Map2Map, but the decoder outputs an output vector instead of a map (with a fully connected layer at the end) and is used only for predicting the 3D position vector and the 3D orientation vector. All architectures have about 1.1M trainable parameters, making them very light compared to most state-of-the-art nets for similar tasks. They are trained for 100 epochs, with AdamW optimizer, using our novel Pytorch-based NGC graph framework. \textbf{Note:} we have a total of 27 edge nets in NGC, each with a unique transformation from input to output representations, totaling about 30M parameters in the entire NGC model.

\noindent \textbf{Unsupervised learning over multiple iterations:}
The unsupervised learning experiments follow the steps of Algorithm \ref{alg:NGC_learning}: during the supervised stage (Step 1), we use the 8k labeled images (train set) to train single EdgeNets to predict one representation from another. Next (Step 2), we create 2-hop paths, from rgb inputs to all representations, by using all the remaining representations as intermediate nodes. 
Then, we evaluate individually each 2-hop path on the 2k images validation set (part of the labeled train set) and add them, in greedy fashion to the NGC graph, based on their individual performance, to form ensembles, as long as the ensemble performance on the validation set keeps improving. This graph construction procedure is very fast, since the validation set is small (2k images). The actual graph structure discovered in this manner (simpler version in Fig. \ref{fig:NGC_Actual_Model}B) is (list of input nodes with edge nets towards the same output node): 1) (rgb, halftone, semantic segmentation, surface normals (camera-view), surface normals (world view)) $\rightarrow$ depth, 2) (rgb, wireframe, surface normals (world-view)) $\rightarrow$ surface normals (camera-view), 3) (rgb, wireframe, surface normals (camera-view), halftone) $\rightarrow$ surface normals (world-view), 4) (rgb, halftone, surface normals (world-view)) $\rightarrow$ semantic segmentation, 5) (rgb, halftone) $\rightarrow$ wireframe 6) (rgb, surface normals (camera-view), surface normals (world-view), semantic segmentation, halftone) $\rightarrow$ absolute 6D pose (3D position + 3D orientation). This gives a total of 27 EdgeNets in our NGC model, each having a unique pair of (input, output) representations. Then we proceed with the unsupervised learning phase (Step 3), for two iterations of learning on test set 1 (iteration 1), then test set 2 (iteration 2). The evaluation is performed on the unseen evaluation set and results reported in Table \ref{table:table_representations_results}. Note that NGC learning generally improves, for all tasks and representations, from one iteration to the next, significantly outperforming the supervised version (trained only on the labeled train set). \\

\noindent \textbf{Comparisons on multi-task learning:}
NGC shares common goals with multi-task and semi-supervised learning. 
Most multi-task methods focus on how to combine the weights of several neural networks \cite{misra2016cross,rosenbaum2017routing,ruder2017overview}. We selected NDDR~\cite{gao2019nddr} and MTL-NAS~\cite{gao2020mtl}, state-of-the-art in 2019 and 2020 on NYU-Depthv2~\cite{Silberman:ECCV12} on two-task learning (semantic segmentation and camera normals) and tested them on our dataset
(Tab.~\ref{table:table_multitask_experiments}).
\\

\noindent \textbf{Comparisons on semi-supervised learning:}
Despite a large body of research on semi-supervised learning, most solutions are tailor-made for specific tasks. 
For comparison, we select CCT \cite{ouali2020semi}, one of the more general approaches with state-of-the-art results on PascalVOC~\cite{everingham2010pascal}. The method is weakly related to ours, as it also exploits consensus by perturbing the output of multiple decoders. Again we test CCT on our dataset, using the same labeled and unlabeled data during semi-supervised learning and report results
in Tab. \ref{table:table_semi_experiments}. Our NGC (with 30M parameters total and no pretraining on other datasets) is smaller and more effective than the CCT model (46M parameters, with pretrained ResNet50 \cite{he2016deep} backbone). We outperform CCT also in relative improvements over the supervised baselines.
\\

\begin{table}
\caption{\label{table:table_ensembles_results} Comparison with baseline ensembles. Results are reported on our evaluation set for two learned representations: semantic segmentation and depth estimation. Results show that our NGC is superior to both types of ensembles on both learned tasks.}
\begin{center}
\begin{tabular}{|l|l|c|}
\hline
Representation & Ensembles &  Metric \\
\hline
 & & mIOU \\
 \cline{2-3}
 & Mixt &  0.4605\\
 \cline{2-3}
Semantic Segmentation & EdgeNets & 0.5248 \\
\cline{2-3}
 & NGC (iteration 2) &  \textbf{0.5258}\\
\hline
& & L1 (meters) \\
 \cline{2-3}
 & Mixt & 3.6608\\
 \cline{2-3}
Depth & EdgeNets & 3.3509 \\
\cline{2-3}
 & NGC (iteration 2) &  \textbf{3.2994}\\
\hline
\end{tabular}
\end{center}
\end{table}

\noindent \textbf{Comparisons to different ensembles of networks:}
Another relevant comparative experiment is against multiple types of vanilla ensembles. We present results on the evaluation set in Table~\ref{table:table_ensembles_results} on semantic segmentation and depth prediction. The number of nets forming the ensembles is the same as the ones in the NGC corresponding to each specific tasks. All nets forming the ensembles were trained using the same setup and training dataset as all our EdgeNets. We consider two types of ensembles: \textbf{1) Mixt ensemble:} we combine different semantic segmentation models, PSPNet~\cite{zhao2017pyramid}, DeepLabV3+~\cite{chen2018encoder} together with our EdgeNet and took the average over the three outputs. We lower the number of parameters of PSPNet and DeepLabV3+ to match the number of EdgeNet (1.1M). We adopt the same procedure for depth prediction as well, for which we also add Unet~\cite{ronneberger2015u} and a larger variant of EdgeNet. \textbf{2) EdgeNets ensemble: } We combined different variants of EdgeNet, by varying the number of parameters (1M, 2M or 10M parameters). We also modified the architecture such that instead of concatenating the features in a Unet-like style, we added them. More details about the architectures are in the appendix.

\section{Conclusions}
We present the neural graph consensus (NGC) model, a novel approach to multi-task semi-supervised learning, which brings together the power of neural networks and that of discrete graphs, by combining many different deep networks into a large neural graph that learns, semi-supervised from mutual consensus among multiple pathways. Our extensive experiments and comparisons to top methods, backed by sound theoretical analysis, clearly prove the effectiveness of the model. Our actual implemented NGC, totaling 30M parameters but incorporating no less than 27 nets, learns to predict semi-supervised, from single images and with top performance, seven different scene interpretations. Our results show that learning from consensual outputs in such large, collaborative multi-task neural graphs, is a powerful direction in unsupervised learning.

\section{Acknowledgements} This work was funded in part by UEFISCDI, under Projects EEA-RO-2018-0496 and
PN-III-P1-1.2-PCCDI-2017-0734. We want to express our sincere gratitude towards Aurelian Marcu and The Center for Advanced Laser Technologies (CETAL) for their generosity and providing us access to GPU computational resources.

\bibliography{egbib}

\clearpage



\twocolumn[
\begin{center}
      {\huge Appendix - Supplementary material}\\
       \vspace{1cm}
\end{center}
]
We provide additional technical and experimental insights and results, to further convince the reader of the effectiveness of our Neural Graph Consensus (NGC) learning approach. We present a video demo of the performance of our NGC model before and after semi-supervised learning, we
give more details about the framework that we developed in order to implement NGC, we discuss how we collected the data and also present the design choices of the deep nets used to create the graph.

\section{Qualitative results in video}

We provide the read a video\footnote{\url{https://www.youtube.com/watch?v=y6sL-lVTPic}}, showing the actual output of the NGC (after unsupervised training) versus the initial networks (EdgeNet iteration 0) for the different prediction tasks addressed.  While our method is based on processing single frames (independently), we present the results in form of a continuous video in order to better assess the quality of the approach in the temporal domain, even though no temporal information was used by our specific NGC implementation - \textbf{each frame is processed independently of the others}. 

The video, running for over 2 minutes (15 FPS, 2076 frames) over a continuous flight within the virtual environment,
shows the output of EdgeNet (iteration 0), the outputs of our NGC (iteration 2) and the regions where NGC is better than EdgeNet. The green pixels depict the regions where NGC (after two iterations of unsupervised learning) is better than EdgeNet (before unsupervised learning), while the red regions show the pixels where they get worse than EdgeNet (iteration 0). The pixels in black are the ones where the two versions, the supervised EdgeNet and the semi-supervised NGC, have the same output (as expected they are more visible in the classification tasks, such as wireframe and semantic segmentation). 

In the video, the camera follows a part of the evaluation set trajectory -- approximately 20\% of the total images. This continuous flight path was never seen during training or semi-supervised learning. We provide snippets of five different tasks: depth, semantic segmentation, surface normals with reference to the camera (C), surface normals with reference to the world (W) and wireframe. 

The improvements in all these different tasks after unsupervised learning (learning from data that has no labels) can be clearly observed. As specified previously, all the learning and prediction is done on a single frame setup. All frames are processed independently with no temporal information used.

\section{Influence of the number of classes}

\begin{figure}
\centering
\includegraphics[scale=0.24,keepaspectratio]{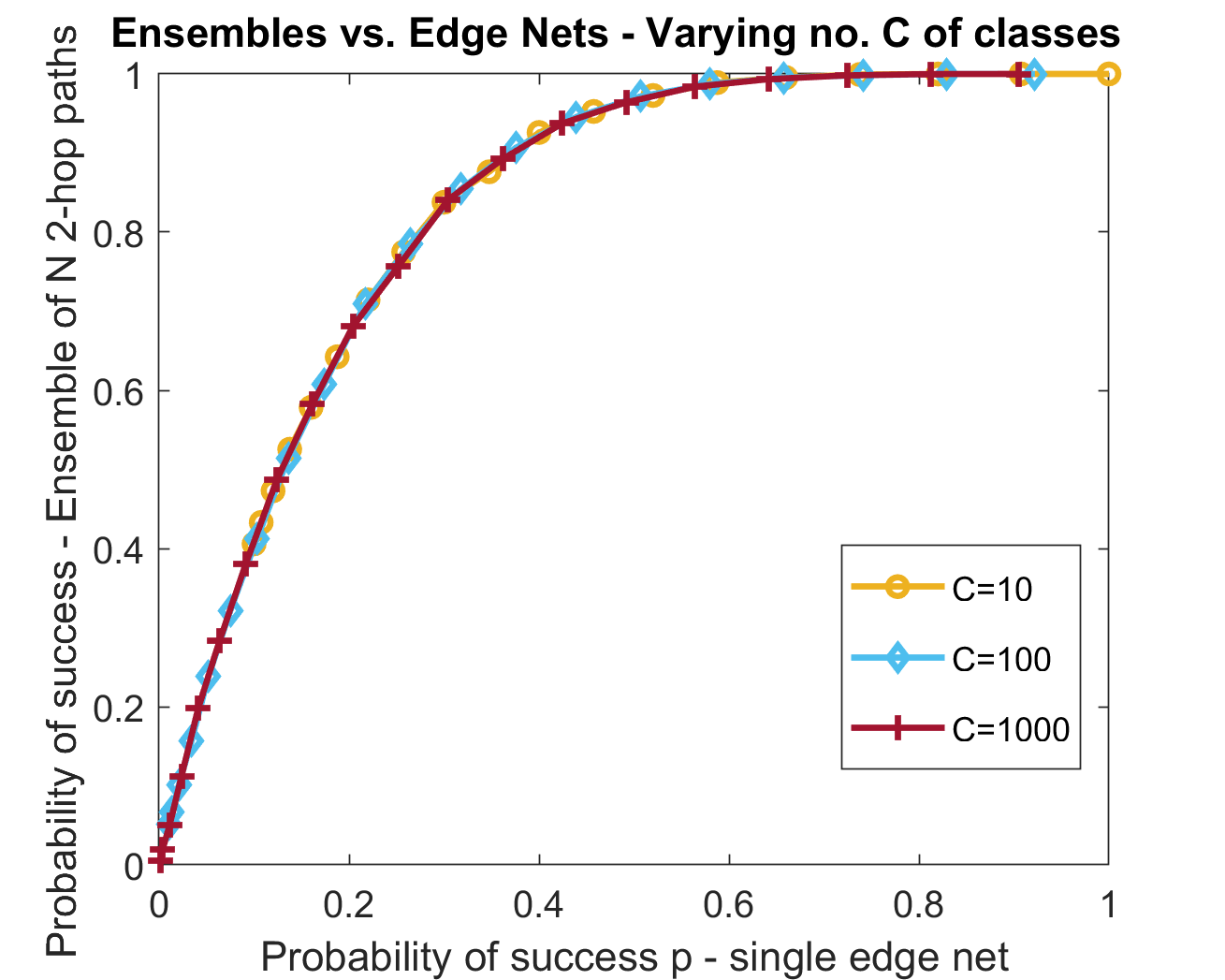}
\caption{\label{fig:varible_classes} Simulation results demonstrate that varying the number of classes (C) does not influence the performance.}
\end{figure}

We provide an addition plot (Figure \ref{fig:varible_classes}) in which we show that varying the number of classes does not influence the overall performance. We provide more details regarding the simulations we have conducted in our Theoretical Analysis section from the main paper, the subsection regarding unsupervised learning in the case of classification. 

\section{Neural Graph Consensus (NGC) framework - additional details}

\subsection{Implementation details}

In order to validate our theoretical analysis and efficiently implement our NGC model, we have developed a general framework that can model arbitrary complex graphs - which we will make publicly available, with source code and all the necessary documentation.

As stated in the paper, the nodes in the graph are interpreted as unique representations (e.g. segmentation, depth, pose, 3D structure), while the edges connect two or more nodes together in order to transform one representation into another. These unique representations are implemented as multi-dimensional tensors, which are generic enough to represent any type of information: 6DoF pose, representations related to images and videos, audio spectrograms or texts (2D) and so forth. Then, each edge must map the input tensor into an output tensor via a transformation function $f:I \rightarrow O$. This function can be static, however, the more interesting case is where this function is represented as a learning model. Particularly, for our experiments, these functions are neural networks that operate on Vector (1D: Position and Orientation) and Map (2D: RGB, Depth, etc.) nodes.

The EdgeNet models can be trained independently (RGB $\rightarrow$ Depth, Pose$\rightarrow$Semantic, etc.), as we do in our experiments, when we pretrain them in a supervised manner on the labeled training set before we start the semi-supervised learning process.

In our NGC framework, graphs can become arbitrarily large, with new deep nets on edges being added or removed as needed. The information flow can follow arbitrary paths and the optimization routes from inputs to outputs can be switched on and off based on how the problem is defined. Since edges are mappings between conceptual nodes, we can use any kind of transformation functions, from static mappings (means, medians) to state-of-the-art neural models (depth or semantic estimators from RGB images). This allows us to reuse existing work, as well as improve other deep nets on edges iteratively, as described in the Theoretical Analysis section.

All the necessary details and instructions for using our framework will be made publicly available.

\begin{table*}
\begin{center}
\centering
\begin{tabular}{|c|c|c|c|c|c|c|c|c|}
\hline
& RGB & Depth & SSeg & Halftone & Surf. Norm. (C) & Surf. Norm. (W) & Wireframe & Pose \\ 
\hline
Depth & \checkmark & & \checkmark & \checkmark & \checkmark & \checkmark & & \\ 
\hline
SSeg & \checkmark & & & \checkmark & & \checkmark & & \\ 
\hline
Halftone & \checkmark & & & & & & & \\ 
\hline
Surf. Norm. (C) & \checkmark & & & & & \checkmark & \checkmark & \\ 
\hline
Surf. Norm. (W) & \checkmark & & & \checkmark & \checkmark & & \checkmark & \\ 
\hline
Wireframe & \checkmark & & & \checkmark & & & & \\ 
\hline
Pose & \checkmark & & \checkmark & \checkmark & \checkmark & & & \\ 
\hline
\end{tabular}
\caption{\label{tab:dependency_table} Dependency matrix for the best graph configuration, used for both iteration 1 and 2. $M(i, j)$ = \checkmark if node $i$ uses intermediate node $j$ during the optimization process. We use the direct link RGB $\rightarrow$ any of the output nodes in the first column, for all of the learned representations. The rest of the dependencies were experimentally determined by optimizing our results on the training set.}
\end{center}
\end{table*}

\subsection{EdgeNet optimization} 

Our EdgeNet of choice is a Unet-like~\cite{ronneberger2015u} architecture with an encoder, bottleneck and decoder. The bottleneck has 6 sequentially dilated convolutions, before applying the decoder.
We have two types of decoders in our graph depending on the output type:
\begin{itemize}
    \item \textbf{Map2Map} (input map, output map), used for semantic segmentation, depth estimation, surface normals prediction (from both camera-view and world-view), wireframe and halftone.
    \item \textbf{Map2Vector} (input map, output vector), used 6DoF pose estimation (for both position and orientation). 
\end{itemize}

Each EdgeNet has $\approx$1.1M parameters. We conducted all our experiments in PyTorch~\cite{NEURIPS2019_9015}. All networks were trained in the same setup, for 100 epochs, using AdamW optimizer, a learning rate of 0.001 and a constant seed for weights initialization. 
\subsection{Building the graph from independent single links} 

Each graph is composed of multiple EdgeNets, which are edges of the graph trained independently and then connected to create the Graph-Ensembles (NGC). In our setup, the maximum depth of the graph is 2, so, for each task T, we either have the direct RGB$\rightarrow$Task or RGB$\rightarrow$Intermediate$\rightarrow$Task. With this in mind, we have defined two types of edges, the ones on the left side of the graph RGB$\rightarrow$Intermediate (first links) and the ones on the right side of the graph Intermediate$\rightarrow$Task (second links). We are now left with the question of how to build the graph from these links in order to create the NGC. For N possible pathways, we have N! possible subgraphs. We chose a greedy approach, based on the training set error of each task alone.

The greedy selection results in all the possible paths for a task sorted in descending order (higher performance first):
\begin{itemize}
    \item RGB$\rightarrow$Task
    \item RGB$\rightarrow$$I_1$$\rightarrow$Task
    \item \ldots
    \item RGB$\rightarrow$$I_n$$\rightarrow$Task
\end{itemize}

Then, we create N possible graphs (by adding one pathway at a time) and compare the error for each possible NGC. Finally, we chose the one that yielded the best results and used it for both iteration 1 and 2. We repeated this process for each task (Depth, Semantic Segmentation, World Normals, Camera Normals, Wireframe and Pose) and reached the final graph configuration presented in Table~\ref{tab:dependency_table}.

\section{Dataset collection}

We used the Town04 virtual environment provided by the Carla simulator~\cite{dosovitskiy2017carla} to capture the images. For the surface normals with respect to the world representation we used UnrealCV~\cite{qiu2016unrealcv}. For the camera normals, we made a custom representation derived from UnrealCV. For optical flow, we adapted the resources from~\cite{fan_jiang_icra19}. The trajectories were also custom - we aimed to simulate the movement of a drone - therefore, noise was added for angles and current camera location.

\section{Comparisons to different networks ensembles}

In this section we give more details regarding the types of ensembles we trained for each of the two tasks: semantic segmentation and depth estimation to further underline the capabilities and advantages brought by our NGC model. For both learned representations we used two types of ensembles:
\begin{itemize}
\item\textbf{Mixt ensemble.} We average the results of multiple state-of-the-art semantic segmentation models. We used variants of PSPNet~\cite{zhao2017pyramid}, DeepLabV3+~\cite{chen2018encoder} and Unet~\cite{ronneberger2015u} in our experiments.
\item\textbf{EdgeNets ensemble.} We average the outputs of multiple variants of EdgeNets which differ in the number of parameters and also the type of features aggregation. The original version of our Unet-like architecture, EdgeNet, uses concatenation for feature aggregation. These operations are replaced with sum in our EdgeNet-Sum variant.
\end{itemize}

For a fair comparison and also in order to better emulate a similar behaviour to our NGC but in a different way, each of the trained ensemble featured the exact number of dependencies for each learned representation. For example, in the case of semantic segmentation, our best subgraph has 3 links and for depth estimation it has 5 links (see Table~\ref{tab:dependency_table} for more details). In the counting we also included the direct link (EdgeNet). Quantitative results for each trained architecture from our ensembles are reported in Table~\ref{tab:quant_sseg} for semantic segmentation and Table~\ref{tab:quant_depth} for depth estimation.

\subsubsection{Semantic Segmentation}

In the case of our Mixt ensemble we reduced the number of parameters of the initial implementations in order to match the one of our EdgeNet ($\approx$1.1M). Compared to the Mixt ensemble, NGC improves by a large margin. One could argue that the total number of parameters for this ensemble is smaller than NGC and this factor could contribute to these results. Therefore, we also experimented with an ensemble of EdgeNets in which the number of parameters was matched with those of our NGC with the same number of links, 2 with different types of architectures and 1 direct link that was also used by NGC. We demonstrated that increasing the number of parameters alone is not sufficient for better performance. Our NGC, learned with multiple representations, is superior to an ensemble with the same number of parameters.

\begin{table}
\begin{tabular}{|l|l|l|l|}
\hline
Ensemble & Architecture & Num. & mIOU\\ 
Type & Type & params & \\
\hline\hline
Mixt & PSPNet & ~1.1M & 0.4145\\
\cline{2-4}
& DeepLabV3+ & ~1.1M & 0.3775\\
\cline{2-4}
& EdgeNet & ~1.1M & 0.5152\\
\hline
& Average & & 0.4605\\
\hline
EdgeNets & EdgeNet & ~1.1M & 0.5152\\
\cline{2-4}
& EdgeNet & ~2.3M & 0.5198\\
\cline{2-4}
& EdgeNet-Sum & ~2.4M & 0.4789\\
\hline
& Average & & 0.5248\\
\hline
& NGC (iter 2) & & \textbf{0.5258}\\
\hline
\end{tabular}
\caption{\label{tab:quant_sseg} Quantitative results for semantic segmentation using multiple types of ensembles. We report mean IOU over all 13 available classes in our dataset, with the exception of the class "person" since it is missing from our evaluation set (higher is better).}
\end{table}

\subsubsection{Depth Estimation}

In the case of depth estimation, results favor our NGC even more since we compared two ensembles, both with higher number of parameters than NGC (Mixt has $\approx$14.1M, EdgeNets has $\approx$28.8M and NGC has $\approx$9.9M params). The advantages brought by NGC in these experiments are even more striking. In both scenarios NGC shows superior performance, whilst using fewer parameters.

\begin{table}
\begin{tabular}{|l|l|l|l|}
\hline
Ensemble & Architecture & Num. & L1 [m]\\ 
Type & Type & params & \\
\hline\hline
Mixt & PSPNet & ~1.1M & 4.4381\\
\cline{2-4}
& DeepLabV3+ & ~1.1M & 4.9629\\
\cline{2-4}
& Unet & ~1.1M & 5.5375\\
\cline{2-4}
& EdgeNet & ~1.1M & 3.9224\\
\cline{2-4}
& EdgeNet & ~10M & 3.4023\\
\hline
& Average & & 3.6608\\
\hline
\hline
EdgeNets & EdgeNet & ~1.1M & 3.9224\\ 
\cline{2-4}
& EdgeNet & ~2.3M & 4.4193\\ 
\cline{2-4}
& EdgeNet & ~10M & 3.4023\\ 
\cline{2-4}
& EdgeNet-Sum & ~2.4M & 4.0692\\ 
\cline{2-4}
& EdgeNet-Sum & ~10M & 3.7437\\ 
\hline
& Average & & 3.3509\\
\hline
& NGC (iter 2) & & \textbf{3.2994}\\
\hline
\end{tabular}
\caption{\label{tab:quant_depth}Quantitative results for depth estimation using multiple types of ensembles. We report L1 error in meters (lower is better) on our evaluation set, unseen during training.}
\end{table}

\section{Making code available}

There is a strong need today for video data with multiple labeled representations. To advance research in the fields of multi-task, semi-supervised, self-supervised and unsupervised learning, we will make our code publicly available for data collection and learning with our NGC. The code, dataset and more information about this project will be available at our project page. \footnote{\url{https://sites.google.com/site/aerialimageunderstanding/semi-supervised-learning-of-multiple-scene-interpretations-by-neural-graph}}

\end{document}